\documentclass[graybox]{svmult}

\usepackage{type1cm}        
\usepackage{makeidx}         
\usepackage{graphicx}        
\usepackage{multicol}        
\usepackage[bottom]{footmisc}

\usepackage{newtxtext}       %
\usepackage{newtxmath}       

\makeindex             

\begin{document}

\title*{Fuzzy Classification of Multi-intent Utterances}
\author{Geetanjali Bihani and Julia Taylor Rayz}
\institute{Geetanjali Bihani \at Purdue University, West Lafayette, USA, \email{gbihani@purdue.edu}
\and Julia Taylor Rayz \at Purdue University, West Lafayette, USA, 
\email{jtaylor1@purdue.edu}}
%
%
\maketitle

\abstract{Current intent classification approaches assign binary intent class memberships to natural language utterances while disregarding the inherent vagueness in language and the corresponding vagueness in intent class boundaries. In this work, we propose a scheme to address the ambiguity in single-intent as well as multi-intent natural language utterances by creating degree memberships over fuzzified intent classes. To our knowledge, this is the first work to address and quantify the impact of the fuzzy nature of natural language utterances over intent category memberships. Additionally, our approach overcomes the sparsity of multi-intent utterance data to train classification models by using a small database of single intent utterances to generate class memberships over multi-intent utterances. We evaluate our approach over two task-oriented dialog datasets, across different fuzzy membership generation techniques and approximate string similarity measures. Our results reveal the impact of lexical overlap between utterances of different intents, and the underlying data distributions, on the fuzzification of intent memberships. Moreover, we evaluate the accuracy of our approach by comparing the defuzzified memberships to their binary counterparts, across different combinations of membership functions and string similarity measures.}

\section{Introduction}
\label{sec:1}

Dialog-based systems have become increasingly ubiquitous, extending their range of conversational ability from open-ended conversations to task-oriented settings. While open-ended dialog systems engage with the user in order to participate in a conversation \cite{gopala2019topical}, task oriented dialog systems focus on completing specific tasks enunciated by the user in the form of utterances, i.e. written or spoken natural language statements \cite{liu2018end}. These utterances describe particular goals as enunciated by the speaker. For example, when a speaker provides an utterance, \textit{`What is the temperature in the room?'}, the associated intent class can be formalized as \textit{`Get Temperature'}. Moreover, a single utterance can not only encode one, but multiple intents. For example, when a speaker provides an utterance, \textit{`What is the room temperature and also play some music.'}, the associated intent classes can be formalized as \textit{`Get temperature'} and \textit{`Play music'}.

In order to ingest, process and respond to a user utterance, the standard dialog system architecture comprises of modules that perform speech recognition, natural language understanding, dialog management and natural language generation \cite{minker2004speech}. The task of disambiguating user utterances is accomplished through intent determination, usually performed within the natural language understanding module. 

Prior works frame intent determination as a classification problem ranging from rule-based template matching \cite{dowding-etal-1993-gemini} to data-driven methods, including statistical and neural network models  \cite{bothe2018conversational, Chen_2016, masumura2018multi, ravuri2016comparative, wang12002combination}. While rule-based approaches guarantee accuracy, they do not account for unseen utterances. Data-driven, in particular statistical approaches show improvement by extracting corpus-based features to perform classification. But their performance is restricted by the quality of training data and respective features. Recently, neural network approaches have been shown to outperform statistical models in terms of classification accuracy, when classifying intents \cite{ravuri2016comparative}.

Although highly accurate, the fundamental assumption in such classification approaches is that the class membership of utterances within intent classes is binary. Hence, an utterance is limited to either a full or null membership within a given intent class. This assumption strays from the actual nature of natural language utterances, where vagueness of linguistic boundaries promote ambiguity in utterances and respective illocutionary forces \cite{lakoff1975hedges}. To our knowledge, no prior works have addressed this limitation, leading to a lack of resources, including text corpora and techniques, to build fuzzy theoretic classification approaches of natural language utterances.

To address this gap, this work builds an approach to create fuzzy membership labels for multi-intent utterances, i.e. utterances that contain multiple intents. This work extends the binary intent membership of an utterance to a degree membership setting, accounting for ambiguity in natural language statements. We evaluate our approach over multiple membership functions, datasets and fuzzy string similarity mapping techniques, to identify the optimal fuzzy membership generation approach for utterances with differing levels of overlap within the same intent as well as across different intents.

\section{Related Work}
\label{sec:2}
With the advent of deep learning, intent classification architectures have evolved to variants of recurrent neural networks \cite{bhargava2013easy, ravuri2016comparative}, word embeddings with convolutional neural networks \cite{hashemi2016query, kim2017two}, gated recurrence units \cite{ravuri2016comparative} and end-to-end memory networks \cite{Chen_2016}. Approaches that include preceding context \cite{bothe2018conversational, lee2016sequential}, task oriented pretraining \cite{paranjape-neubig-2019-contextualized, wu2020tod} and emerging intent identification \cite{xia2018zero} during model training, have also shown to achieve state of the art performance. 

Multi-intent classifiers are usually trained on multi-intent training data. These models rely primarily on large corpora for training and testing the model. This approach is not efficient when considering the lack of multi-intent utterance data resources. Thus, instead of relying on multi-intent data resources to learn class memberships, we formulate a process to utilize single intent utterances to perform multi-intent classification.

Current methods of intent classification, including single intent and multi-intent classification, limit themselves to a binary set membership of utterances within any given intent. These methods are built on datasets that do not include ambiguous natural language utterances, portraying a limited view of reality. For example, let's consider an ambiguous utterance, ``I want to open an account.''. In the banking domain, this utterance can mean that the user wants to open a banking account. But, the utterance doesn't specify whether it should be a `checking' account or a `savings' account \cite{dhole2020resolving}, and the corresponding intent remains vague. Moreover, if the utterance is not spoken within the banking domain, it can also refer to opening an online account, e.g. a social media account. This shows that an utterance can be ambiguous leading to vagueness in mapping to specific intents. This ambiguity in natural language utterances can lead to errors in understanding and downgrading of the entire dialog system’s performance \cite{li2017investigation}. In the given example,this can lead to incorrectly identifying the intent to be `Open bank account', when meant as `Open social media account' or vice versa. To improve language understanding, recent papers have also focused on identifying emerging intents where labelled utterances are absent \cite{xia2018zero}, asking clarification questions to resolve intent ambiguities \cite{dhole2020resolving}, etc. 

Prior approaches exploring imprecision in text utilize fuzzy logic to assign membership degrees within semantic categories at word level \cite{andreevskaia-bergler-2006-mining, subasic2001affect} as well as sentence level \cite{fu-wang-2010-chinese}. Mostly, the consideration of utilizing fuzzy sets to address imprecision and centrality has been limited to sentiment classification tasks. Fuzzy rule based intent classification has been limited to the use of fuzzy likelihood, and not extended to include fuzzy set membership \cite{vu2020fuzzy}. Thus, this work builds a framework that utilizes imprecision in user utterances to assign degree memberships within intents, allowing not only multiple classes but also multiple degrees of memberships..


\section{Utterance Level Fuzzy Memberships}
\label{sec:3}

Task-oriented dialog systems primarily employ ‘directive’ style utterances, spoken in the form of commands or requests. These utterances are associated to respective intents or 'illocutionary points', as described by prior works on Speech Act Theory \cite{austin1975things, searle1985expression}. In his work on hedges and meaning criteria, Lakoff described the concept of fuzziness in natural language, stating that natural language utterances can be \textit{`true to a certain extent and false to a certain extent'} \cite{lakoff1975hedges}. This concept has been addressed in further detail in \cite{taylor2011understanding}, where ambiguity in natural language sentences is addressed using a fuzzy theoretic approach, further discussing the assignment of multiple degrees of membership to a sentence, across different `dimensions'. These dimensions include but are not limited to `acceptability', `appropriateness', `relevance', `saliency' and more. 

In a similar vein, vagueness in natural language utterances can be mapped to imprecise intent classes through fuzzy intent memberships, using knowledge-based and data-driven approaches. These approaches vary in terms of interpretability and adaptability. Knowledge-based approaches allow better model interpretability, but lead to highly parameterized rules as the number of inputs grow. This is attributed to the flat structure of the rule base. On the other hand, data-driven approaches allow rule adaptability, but non-trivial interactions between generated rules and complex aggregation schemes affect their interpretability \cite{hullermeier2015knowledge}. Thus, this work presents two complementary rule generation approaches and evaluates their applicability in fuzzy intent membership generation for natural language utterances.

\subsection{Membership Functions}
\label{sec:mem_func}

In order to generate fuzzy intent class memberships for utterances, we utilize softmax scores generated from neural net classifiers. When considered as class probabilities, softmax scores showcase lack of representation of decision uncertainty, as well as overconfidence in incorrect predictions \cite{joo2020being}. To overcome these limitations in this work, the softmax scores for each intent class are interpreted as a distribution of membership values of natural language utterances within the given intent class. This distribution is utilized to generate memberships of utterances within intents. 

The degree of membership of an utterance within every intent class is expanded from binary to fuzzy by dividing it into three fuzzy sets, i.e. $low$, $medium$ and $high$. The fuzzy sets $low$ and $high$ are interpreted as open (edge) sets. Both sets are mapped using sigmoid functions, where $low$ is a decreasing sigmoid, while $high$ is an increasing sigmoid. The choice of membership functions is one of the most difficult tasks while developing a fuzzy expert system \cite{martine2002linguistic}. In this work, since we deal with utterances comprising of multiple words and modifiers, we aim to simplify the task of membership function generation. Thus, our choice of membership functions is driven by minimal parameterization and we employ gaussian and sigmoid functions as fuzzy membership functions representing fuzzy sets. The shape of the sigmoid function is governed by the parameters $a$ and $c$, where $a$ is the width of transition of the sigmoid curve, and $b$ is the center of the transition. The $medium$ fuzzy set is mapped using a gaussian membership function, governed by the parameters $c$ and $\sigma$, corresponding to distribution mean and standard deviation respectively. The membership functions are described below.  

\begin{equation}
S_{low}(x ; a, b)=\frac{1}{1+e^{a(x-b)}}
\end{equation}

\begin{equation}
S_{high}(x ; a, b)=\frac{1}{1+e^{-a(x-b)}}
\end{equation}

\begin{equation}
S_{medium}(x ; \sigma, c)=e^{\frac{-(x-c)^{2}}{2 \sigma^{2}}}
\end{equation}

The parameters of these membership functions are generated using two approaches, namely the knowledge-based approach and the data-driven approach.

\subsection{Parameter Generation}

\subsubsection{Knowledge Based}

\begin{figure}[t]
\sidecaption[t]
\includegraphics[scale=0.28]{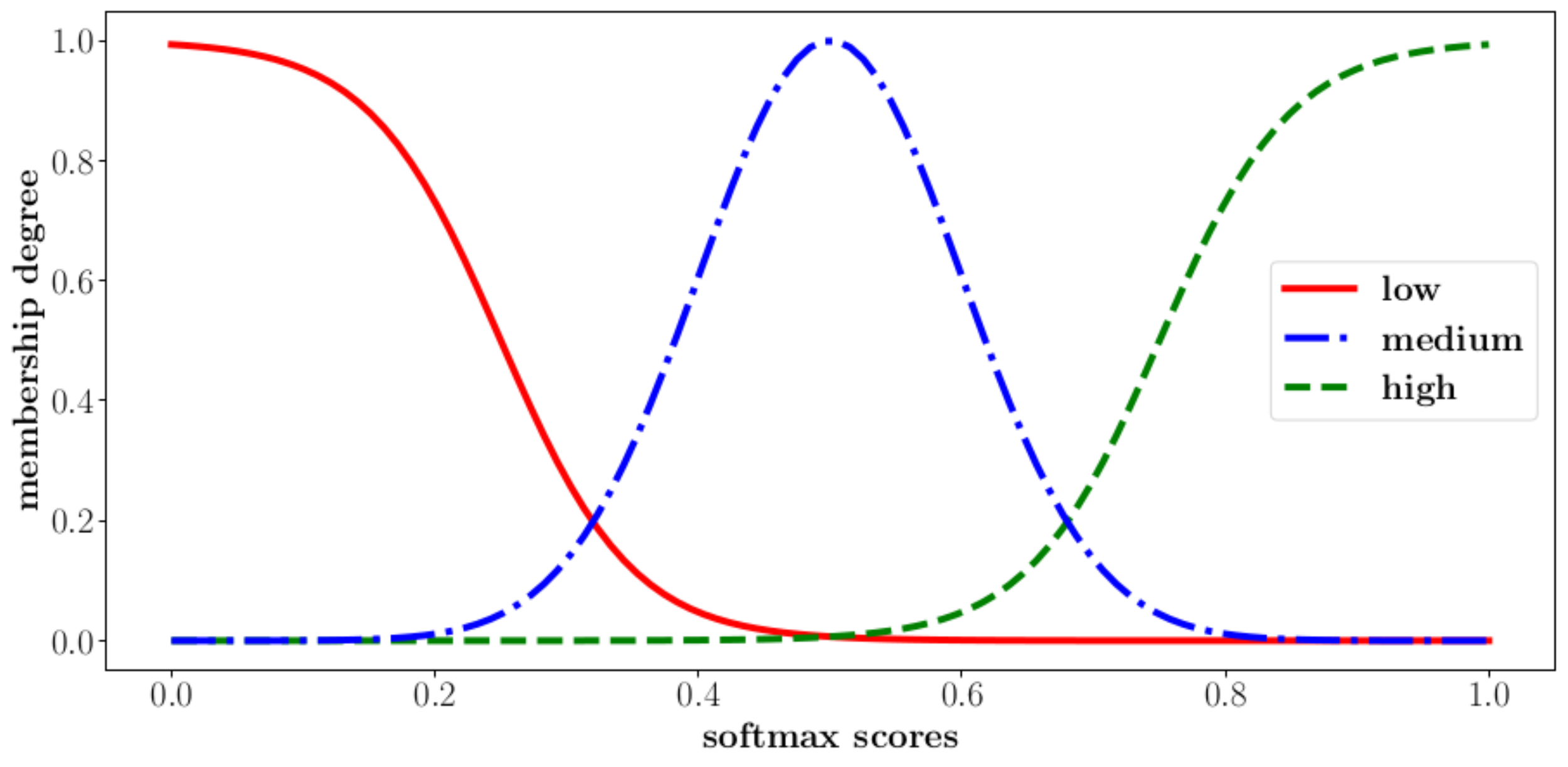}
%
%
\caption{Knowledge-based Membership}
\label{fig:1}       
\end{figure}

Since we are generating membership functions using softmax scores, it is important to understand the realization of these scores into fuzzy functions. Softmax scores are bounded within the range $[0,1]$ and can be interpreted as posterior probabilities $P(\omega|u)$ or $x$, i.e. probability of class $\omega$ given an input utterance $u$. To convert these scores into fuzzy membership sets, knowledge-based parameters are defined in Eq. (4) and (5). Since, $x\in[0, 0.5]$ suggests a lower chance of an utterance being labelled as an intent, the value of $a_{low}$ is chosen to limit the spread of $S_{low}$ membership function within the range $x\in [0, 0.5]$. With the same intuition, the value of $a_{high}$ is chosen to limit the spread of $S_{high}$ membership function within the range $x\in [0.5, 1]$. The parameters $b_{low}$ and $b_{high}$ control the centers of transition for respective membership functions $S_{low}$ and $S_{high}$. Thus, $b_{low}=0.25$ ensures that the center of transition for the membership function $S_{low}$ stays in the middle of the range $[0,0.5]$. Similarly, $b_{high}=0.75$ fixes the center of transition for $S_{high}$ in the middle of the range $[0.5,1]$. The values of $c_{medium}$ is chosen to map the highest degree of $medium$ membership at $x=0.5$, gradually decreasing as $x \to 0$ or $x \to 1$. To limit the spread of the $medium$ membership function within the range of $x\in[0.25,0.75]$, $\sigma_{medium}$ is kept at $0.1$.

\begin{equation}
a_{low}=a_{high}=\frac{k}{0.5}; \> \> b_{low}=0.25; \> \>  b_{high}=0.75
\end{equation}

\begin{equation}
c_{medium}=0.5; \sigma_{medium}=0.1
\end{equation}

These parameters map the softmax scores to corresponding membership degrees within fuzzy sets $low$, $medium$ and $high$, using the membership functions described in Sect.~\ref{sec:mem_func}. These knowledge-based membership functions are visualized in Fig. \ref{fig:1}.

\subsubsection{Data Driven}

\begin{figure}[t]
\sidecaption[t]
\includegraphics[scale=0.75]{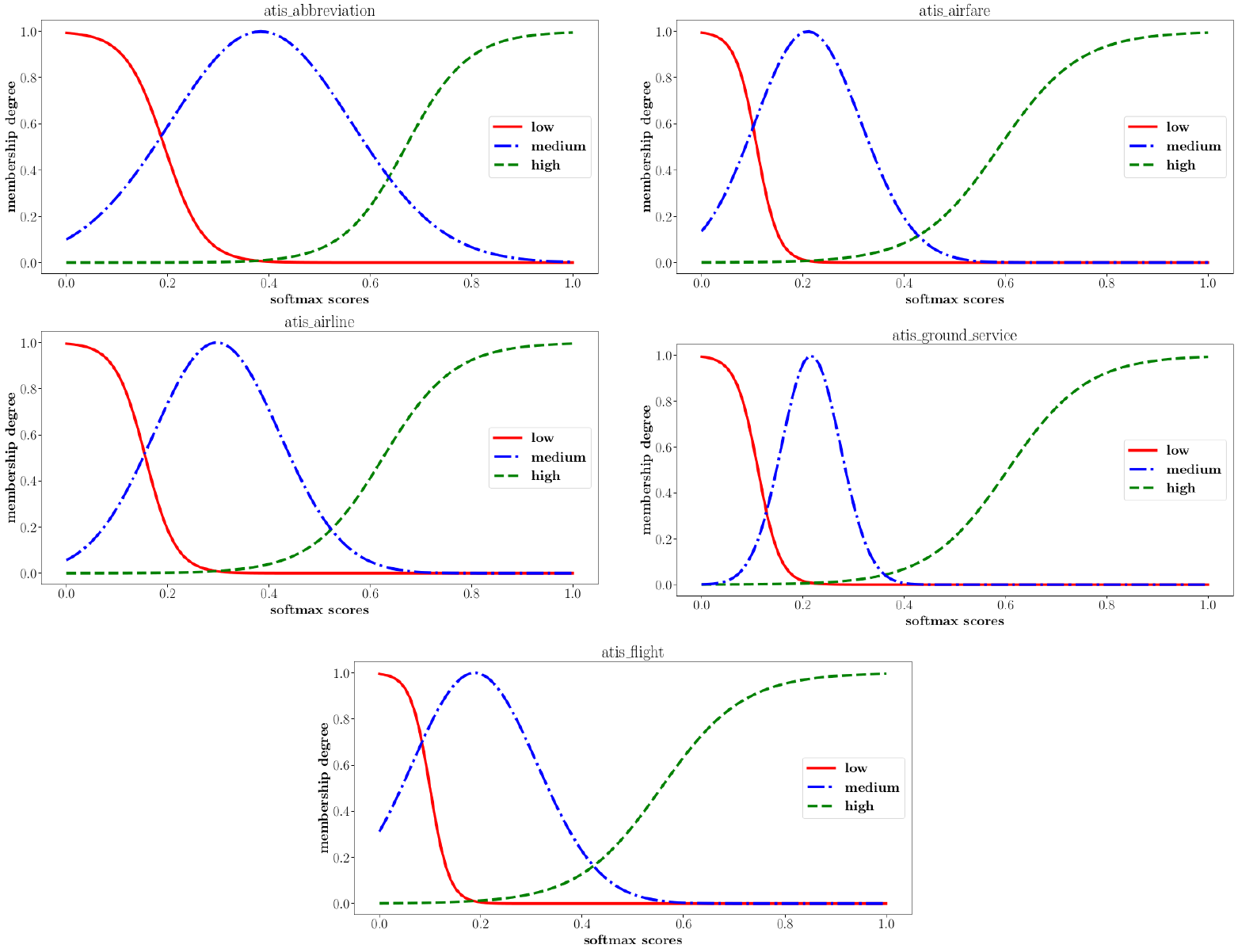}
%
%
\caption{Data-driven Membership}
\label{fig:2}       
\end{figure}

The data-driven approach utilizes utterance labels to derive membership function parameters from softmax scores. For every intent class (such as \textit{Ground Service, Flight, Airfare}, etc.), the distribution of softmax scores $x$ is utilized to calculate the membership function parameters as shown from Eq.(6)-(9), forming different membership functions for every intent class. Here, $x_{L}$, $x_{M}$ and $x_{H}$ are mean softmax scores for respective membership sets $low$, $medium$ and $high$, used to derive their membership functions. Fig. \ref{fig:2} shows an example of the data-driven membership generation approach for five different intent classes retrieved from the ATIS dataset  \cite{hemphill1990atis}. The dataset is described in further detail in Sect.~\ref{ssec:1}

\begin{equation}
x_{L}=\mathbb{E}[x: u\notin I_{C}]; \> \> x_{H}=\mathbb{E}[x: u\in I_{C}]; \> \> x_{M}=\mathbb{E}[x \in [x_{L}, x_{H}]]
\end{equation}

\begin{equation}
a_{low}=\frac{k}{x_{M}}; \> \> b_{low}=\frac{x_{L} + x_{M}}{2}
\end{equation}

\begin{equation}
a_{high}=\frac{k}{1- x_{M}}; \> \> b_{high}=\frac{x_{M} + x_{H}}{2}
\end{equation}

\begin{equation}
c_{medium}=\mathbb{E}[x \in [x_{L}, x_{H}]]; \> \> \sigma_{medium}=\mathbb{SD}[x \in [x_{L}, x_{H}]]
\end{equation}

\section{Single intent to Multi intent utterances}
\label{sec:4}
As their name suggests, multi-intent utterances contain multiple intents, enunciated using phrases or sub-sentences pointing towards individual intents. As discussed in Sect.~\ref{sec:2}, current methods of multi-intent classification are entirely data dependent. In other words, the learning in these systems is limited to the multi-intent data used to train the classifier. This has led to the development of models that perform well for the restricted number of multi-intent datasets available to the NLP community. Thus, we wanted to create a method that is not limited by the sparsity of multi-intent utterance data and can utilize the information in single intent utterances to learn multi-intent utterance classification. In this regard, prior works have considered multi-intent utterances as a combination of single intent utterances, formulating the problem of splitting the combined utterance into sub-parts \cite{kim2017two}. A serious limitation in these approaches is the rule-based nature of splitting, e.g., dividing sentences based on the position of conjunctions or punctuation marks.  To address these limitations, this work performs sentence splitting, not using a rule base, but by approximately matching multi-intent utterances to candidate single intent utterances. Fuzzy IR techniques have shown effectiveness in matching partially related text data \cite{alzahrani2010fuzzy}. Thus, the problem becomes matching fragments of a multi-intent utterance to single intent utterances. Given a knowledge base containing only single intent utterances, we map each multi-intent utterance to $n$ candidate single intent utterances. This approach ensures that the inherently ambiguous nature of natural language utterances is addressed, and new or unseen multi-intent utterances are attended to. This work utilizes fuzzy set similarity measures to conduct partial matching of multi-intent utterances to single intent utterances.

\begin{itemize}
\item{\textbf{Jaccard Similarity:} Given two strings $U_{A}$ and $U_{B}$, the words forming each string can be considered as sets $A$ and $B$. Then, jaccard similarity between the strings is calculated by dividing the number of elements in the intersection of A and B ($|A \cap B|$) and the union of A and B ($|A \cup B|$).  

\begin{equation}
\mathbf{S}_{jaccard}(A, B)=\frac{|A \cap B|}{|A \cup B|}
\end{equation}}

\item{\textbf{Cosine Similarity:} Given two strings $U_{A}$ and $U_{B}$,  cosine similarity between them is calculated by converting each string into TF-IDF vectors \cite{sammut2010tf} and measuring the cosine of the angle between these vectors using the following equation.
\begin{equation}
\mathbf{S}_{cos}(A, B)=\frac{A \cdot B}{\|A\|\|B\|}
\end{equation}}

\item{\textbf{Partial Ratio:} Given two strings $U_{A}$ and $U_{B}$, where $U_{A}$  is the shorter string, partial ratio finds the best matching substring in $U_{B}$ for $U_{A}$. These strings are tokenized to form word token sequences $A$ and $Y \in B$. This measure is calculated using the following equation, where $lev(A, Y)$ represents the levenshtein distance \cite{levenshtein1966binary} between the tokenized word set $A$ from $U_{A}$ and $Y$ from $U_{B}$.

\begin{equation}
\mathbf{S}_{partial}(A, B) = \max_{Y \subset B, \lvert Y \rvert = \lvert A \rvert }\left[\frac{ \lvert A \rvert  +  \lvert Y \rvert  - lev(A, Y)}{ \lvert A \rvert  +  \lvert Y \rvert }\right]
\end{equation}
}

\item{\textbf{Token Set Ratio:} This is a variant of the aggregation of inclusion measure (Cross, 1993), where the edit distance is utilized and the aggregation is performed using the $max$ function. Given two strings $U_{A}$ and $U_{B}$, they are tokenized to form word token sequences $A$ and $B$. The levenshtein ratio ($S_{ratio}$) is measured between the sorted intersections and remainders of these sequences to calculate token set ratio, as given below. 

\begin{equation}
t_{0} = (A \land B)^{s} ;  t_{1} = t_{0} \lor (A \land (\overline{A \land B}))^{s} ; t_{2} = t_{0} \lor (B \land (\overline{A \land B}))^{s}
\end{equation}

\begin{equation}
S_{ratio}(x, y) = \frac{ \lvert X \rvert  +  \lvert Y \rvert  - lev(X,Y)}{ \lvert X \rvert  +  \lvert Y \rvert }
\end{equation}

\begin{equation}
\mathbf{S}_{tsr}(A, B)=\max_{\substack{t_{i} \neq t_{j} \\ t_{i} \in \{t_{0}, t_{1}\} \\  t_{j} \in \{t_{1}, t_{2}\} }}\left[\mathrm{S}_{ratio}(t_{i}, t_{j})\right]
\end{equation}}

\end{itemize}

\section{Fuzzy Membership Aggregation and Defuzzification}
\label{sec:5}

\begin{figure}[t]
\sidecaption[t]
\includegraphics[scale=0.6]{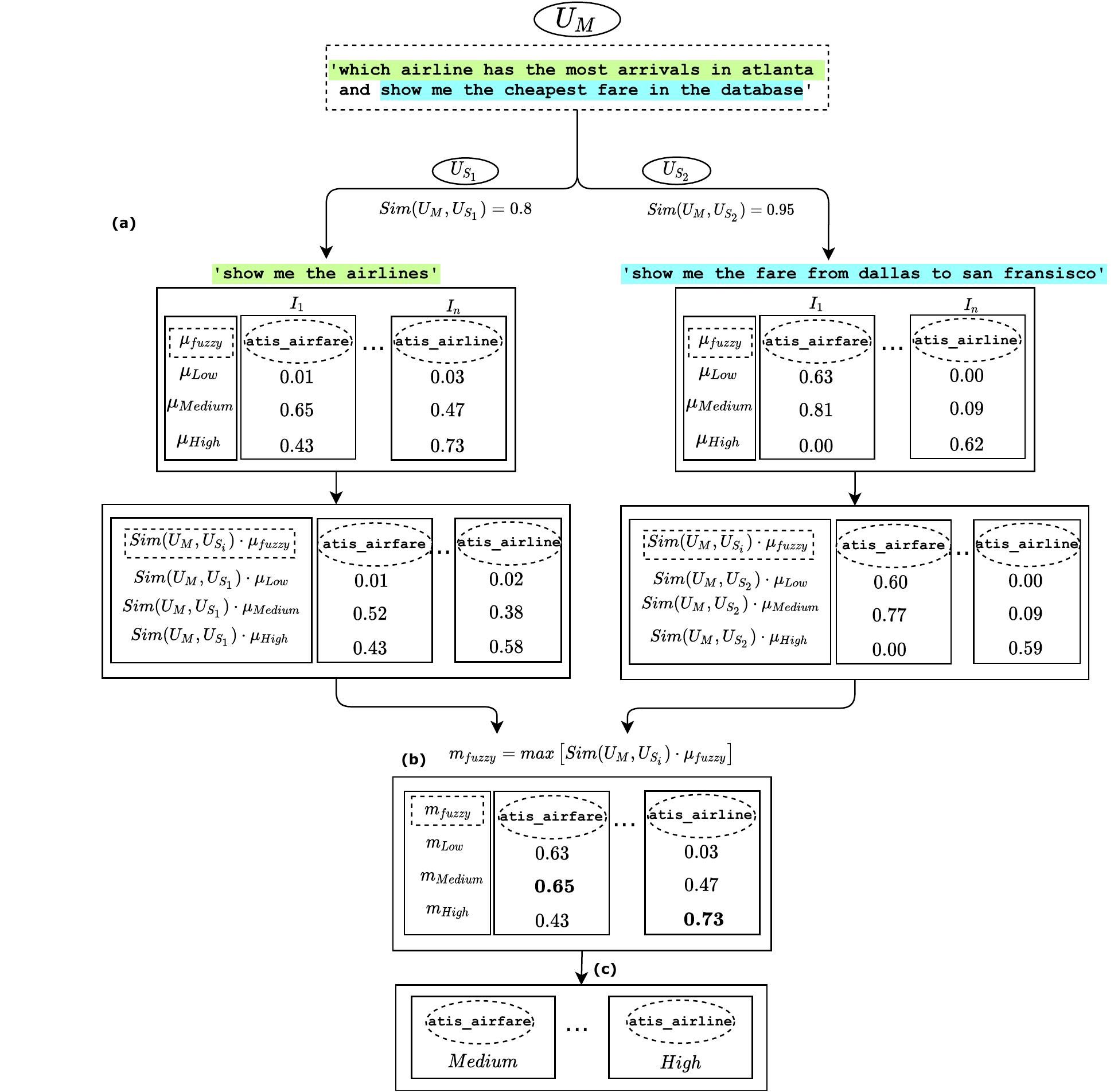}
%
%
\caption{Fuzzy Membership Aggregation Module}
\label{fig:3}       
\end{figure}

Using approximate matching, a multi-intent utterance ($U_{M})$ is mapped to $n$ most similar single intent utterances ($U_{S_1}...U_{S_n} \in S$). For every matched single intent utterance $U_{S_i}$, we get a separate set of fuzzy intent degree memberships. For example, as shown in Fig. \ref{fig:3} (a), $U_{M}$ is associated with $U_{S_1}$ and $U_{S_2}$, where for intent class `\textit{Airfare}', $\mu_{high}(U_{S_1}) = 0.43$ and $\mu_{high}(U_{S_2}) = 0$. Thus, $U_{M}$ has 2 different degrees of `$high$' membership within `\textit{Airfare}', which need to be coalesced to form a singular value. This is done by calculating the similarity-membership product for all $U_{S_{i}} \in S$ and picking the maximum value, as shown in Eq. (16).

\begin{equation}
\mathrm{\mu_{fuzzy}}(U_{M}, I_{c})=\max_{U_{S_{i}} \in S}[sim(U_{M}, U_{S_{i}})\cdot\mathrm{\mu_{fuzzy}}(U_{S_{i}}, I_{c})]
\end{equation}

In order to calculate the efficiency of our approach, we defuzzify intent classes and compare the defuzzified intents with actual intent labels. To defuzzify intent memberships, we utilize the piece-wise function as given in Eq. (17).

\begin{equation}
\mathrm{\mu}(U_{M}, I_{c})=
 \begin{cases} 
      \mu_{low} & \mu_{low}>0.5, \mu_{medium}\leq 0.5, \mu_{high}\leq 0.5 \\
      \mu_{medium} & \mu_{medium}> 0.5, \mu_{high}\leq 0.5 \\
      \mu_{high} & \mu_{high}> 0.5 \\ 
   \end{cases}
\end{equation}

\section{Experiments}
\label{sec:6}
The efficiency of the proposed fuzzy intent classification approach is assessed over two fuzzy membership generation approaches (Sec. \ref{sec:3}), four different approximate string similarity measures (Sec. \ref{sec:4}) and two different multi-intent classification datasets (Sec. \ref{ssec:1}). 

\subsection{Setup}
\begin{figure}[t]
\sidecaption[t]
\includegraphics[scale=0.6]{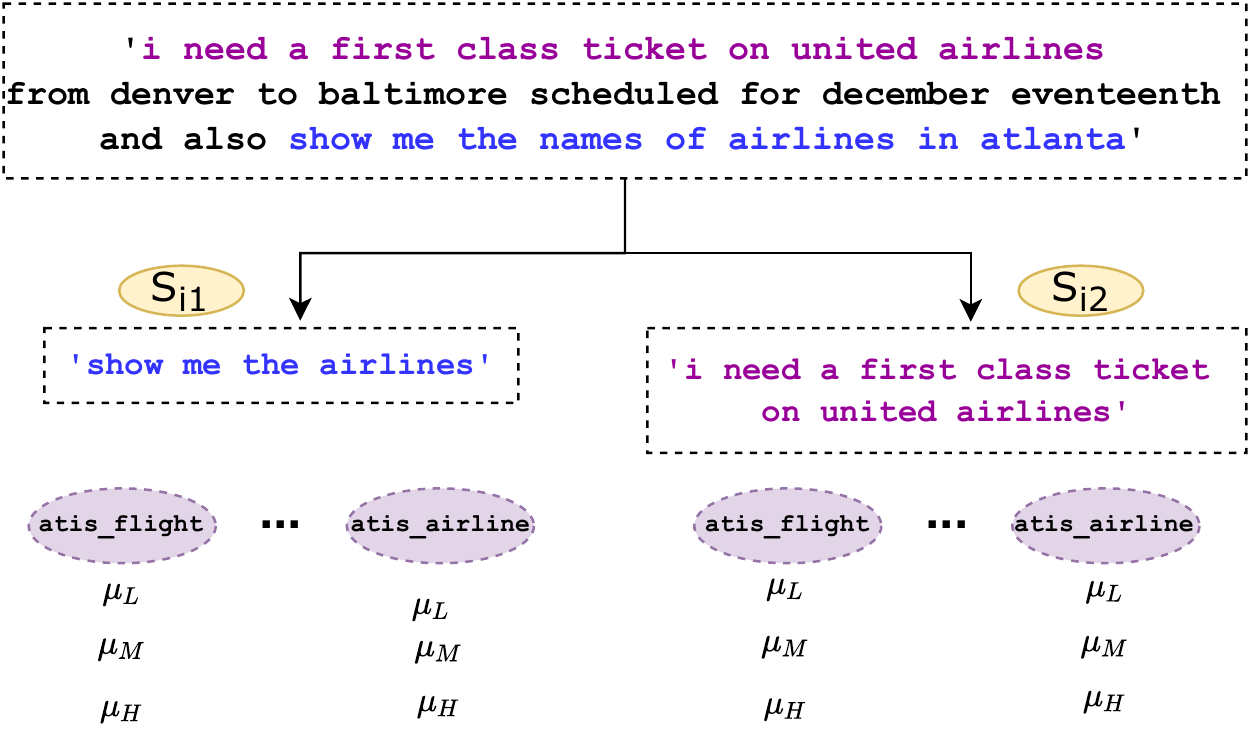}
%
%
\caption{Overall Architecture}
\label{fig:4}       
\end{figure}

The system is divided into three modules, i.e. fuzzy membership generation module, utterance mapping module and fuzzy membership aggregation module. This encapsulates the typical fuzzy inference process, involving fuzzification, implication and defuzzification. The process starts with the fuzzy membership generation module, where softmax scores are used to create fuzzy memberships of SI utterances within intent classes. The utterance mapping module maps every incoming MI utterance to the 3 most similar single intent utterances and respective memberships within the single intent utterance database. This database contains randomly $k$ sampled SI utterances ($k = 1000$). The fuzzy membership aggregation module outputs an overall membership of MI utterance within each fuzzy intent class. The architecture is described in Fig. \ref{fig:4}.

%

\subsection{Data} \label{ssec:1}
This approach is assessed over two widely used task oriented intent classification datasets, ATIS (Airline Travel Information System) and SNIPS. The datasets used are the single intent as well as multi intent versions provided by \cite{qin-etal-2020-agif}. The single intent versions are utilized to create the membership model, and the performance of the proposed approach is validated over 5 iterations of $k=1000$ randomly sampled multi-intent utterances from multi-intent utterance datasets (MixATIS and MixSNIPS). The original ATIS dataset has an unbalanced distribution of intents, with $79.1\%$ of the utterances classified under the same intent class (i.e. \textit{Flight}). Thus, this work utilizes a modified version of the dataset, randomly sampling $\sim200$ instances of each class, and limiting the analysis to the 5 most populated intent classes, in order to create a more balanced dataset. The modified ATIS dataset has 1,066 utterances and 5 intents. The SNIPS dataset is considerably larger, containing 14,484 utterances and 7 intents. Since, the latter has a balanced intent distribution, no modifications are made.



\subsection{Training Details}
An LSTM-based neural network classifier is used to generate softmax scores across intents for single intent utterances. The architecture comprises of a $100$ dimensional embedding layer, a dropout layer, an LSTM layer and a dense softmax classification layer, with a learning rate of $\eta=0.001$. 
The ATIS model is trained with a batch size of $20$, while the SNIPS model is trained with a batch size of $50$. For both the datasets, the models are trained over $5$ epochs. 

\subsection{Results}
\begin{figure}[b]
\sidecaption[t]
\includegraphics[scale=0.7]{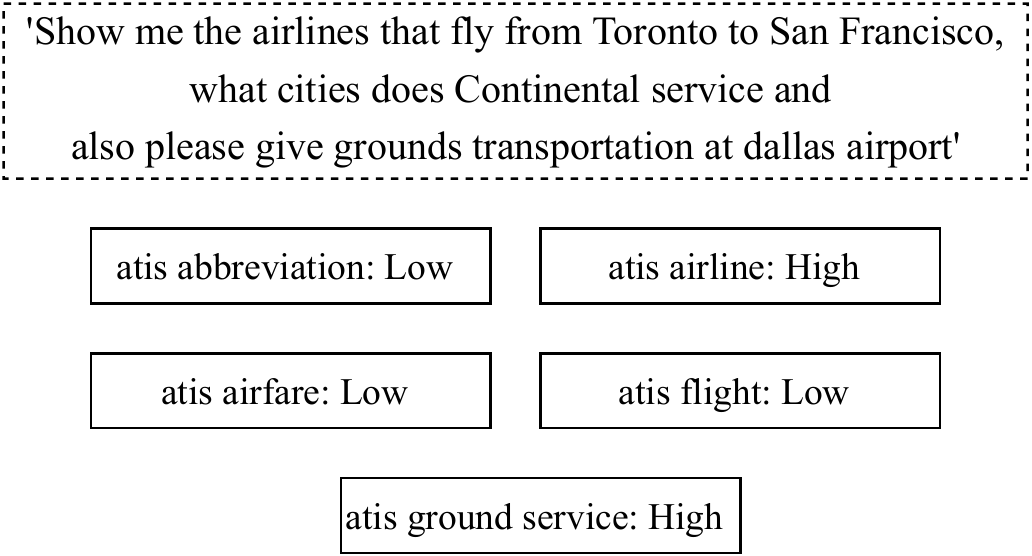}
%
%
\caption{Fuzzy intent memberships for multi-intent utterance}
\label{fig:5}       
\end{figure}

%
The generated fuzzy memberships are assessed by comparing how much they emulate their binary counterparts. Thus, if a multi-intent utterance $U_{M}$ has binary membership of $1$ in an intent class $I_{C}$, a mapping to fuzzy membership of `high' is considered accurate. Likewise, a binary membership of $0$ mapped to fuzzy membership `low' is considered accurate. Thus, higher accuracy signifies that the system creates more binary labels, while lower accuracy portrays more fuzziness. For example, an accuracy of $70\%$ signifies that $30\%$ of the utterances are assigned `medium' memberships, not traceable to their binary counterparts. An example is shown in Fig. \ref{fig:5}, where a multi-intent utterance is assigned fuzzy memberships over intent classes.

\subsubsection{Knowledge-based or Data-driven}
The results in Table \ref{tab:1} show that the accuracy of membership assignment is less variable across datasets when using the data-driven approach. This is because the $\alpha$ cuts for fuzzy memberships are derived from the softmax distribution. As a result, a consistent proportion of utterances is assigned `low', `medium' and `high' membership, for both the datasets, leading to consistency in reported accuracy.

Unlike the data-driven approach, the knowledge-based approach shows highly variable results, showing low accuracy over ATIS, but high accuracy over SNIPS. Note, that the knowledge-based approach derives $\alpha$ cuts for fuzzy memberships using rules, dividing the softmax scores into three groups using constants, instead of variables. These constants do not take the softmax score distributions into account while learning memberships. As a result, the variability in the softmax score distributions of ATIS and SNIPS leads to variability in memberships assigned, leading to highly variable results. 

The softmax score distribution for SNIPS is more bimodal compared to ATIS. This bimodality is due to the softmax scores assigned to utterances being very high (approach $1$), or very low (approaching $0$). This results in the knowledge-based approach tagging a significantly higher proportion of utterances as having `low' or `high' memberships, leading to higher accuracy than ATIS. 

Amongst the two approaches, the data-driven approach is more flexible, creating adaptive memberships over intents for utterances. This flexibility also leads to consistency in emulating binary memberships. On the other hand, the knowledge-based approach is more affected by variations in softmax score distributions, as compared to the data-driven approach.

\begin{table}[!t]
\caption{Fuzzy Classification Accuracy}
\label{tab:1}       
%
%
\begin{tabular}{p{2cm}p{2.8cm}p{2.8cm}p{2.8cm}}
\hline\noalign{\smallskip}
Data & Fuzzy Sim. Metric & Knowledge-based & Data driven              \\
\noalign{\smallskip}\svhline\noalign{\smallskip}
        & Jaccard                 & 57.03 ± 0.016          & 66.11 ± 0.014 \\
        & Cosine                  & 57.98 ± 0.015          & 67.36 ± 0.001 \\
ATIS    & Partial                 & 62.51 ± 0.002          & 65.60 ± 0.002 \\
        & Token Set               & 62.22 ± 0.002          & \textbf{70.18 ± 0.002} \\
        &                         &                        &               \\
       & Jaccard                 & 89.11 ± 0.001          & 71.76 ± 0.001 \\
        & Cosine                  & 89.40 ± 0.001          & 71.72 ± 0.001 \\
SNIPS    & Partial                 & 91.33 ± 0.001          & 71.77 ± 0.001 \\
        & Token Set               & \textbf{91.52 ± 0.000 }         &  72.00 ± 0.001 \\
        
\noalign{\smallskip}\hline\noalign{\smallskip}
\end{tabular}

$^a$ Standard error reported across 5 iterations
\end{table}
\subsubsection{Comparison across Fuzzy String similarity metrics}
In this work, we match strings using lexical similarity between a given multi intent utterance and candidate single intent utterances. As seen from Table \ref{tab:1}, the fuzzy similarity measures show higher variability in performance over ATIS ($\Delta = 4.78$) as compared to SNIPS ($\Delta = 1.35$). This difference is attributable to the difference in lexical overlap between utterances of different intents, which impacts the mapping accuracy of multi-intent utterances to nearest fuzzified single intents. Higher overlap between utterances across intents leads to lower mapping accuracy and vice versa. ATIS contains more lexically similar utterances (homogeneous) across different intents, where phrases such as \textit{`show me'} and \textit{`from'} repeat across intents. On the other hand, SNIPS contains more dissimilar (heterogeneous) utterances, with lesser overlap between utterances of different intents. Examples of both the cases are given below.

\begin{description}[ATIS]
\item[\textbf{ATIS}]
\item[`atis flight':]{\textit{`\textbf{show me flights} to philadelphia coming \textbf{from} Baltimore}'}
\item[`atis airline':]{\textit{`\textbf{show me} airlines with \textbf{flights} \textbf{from} Denver'}}
\end{description}

\begin{description}[SNIPS]
\item[\textbf{SNIPS}]
\item[`PlayMusic':]{\textit{`\textbf{play} say a word by la india'}}
\item[`RateBook':]{\textit{`\textbf{rate} this series 0 of 6 stars'}}
\end{description}

Overall, token set ratio yields highest accuracy, showing that matching substrings by accounting for similarity between intersections as well as remainders produces the best results. Moreover, this measure is least affected by the lexical overlap between utterances of different intents.

\begin{description}[ATIS]
\item[\textbf{ATIS}]
\item[(a) \textit{`Minneapolis to Phoenix on Monday'}]
\item[(b) \textit{`San Francisco to Denver'}]
\end{description}

\begin{description}[SNIPS]
\item[\textbf{SNIPS}]
\item[(c) \textit{`Find the path to power'}]
\item[(d) \textit{`My idea of fun is a book that should get 2 stars'}]
\end{description}

Moreover, ambiguity in single intent utterances leads to uncertainty in intent labelling. This can be observed from the examples above, where (a) and (b) show that even though utterances have provided information in the form of naming sources and/or destinations, the attached illocutionary force or intent remains vague. It is not clear whether the underlying intent is to get a list of airlines or flights or both. Similarly, (c) and (d) are vague utterances, where ambiguity is amplified due to multiplicity in meaning. Here, (c) can be considered as a statement to find some document called \textit{'The Path to Power'} or a statement asking the listener to find the path to power, which will be an unclassified intent, if considering SNIPS intent classes. Similarly, an utterance worded like (d) either suggests that a book called `My Idea of Fun' be given $2$ stars, or that the speaker's idea of fun is a book that should get $2$ stars. These are some of the few examples extracted from ATIS and SNIPS datasets that show that ambiguity in natural language utterances affects the mapping of associated intent class(es).

\section{Conclusion}
\label{sec:7}
 Natural language utterances are seldom precise, containing some form of vagueness. Current research on intent classification, including approaches as well as corpora, are limited to a one-dimensional view, where utterances are treated as atomic inputs, with binary memberships within intent classes. This paper proposes a framework towards fuzzy intent classification for unseen multi-intent utterances, without the need for the existence of prior multi-intent utterance data to learn intent memberships. This framework is assessed over different fuzzy membership generation techniques, fuzzy string similarity measures and different datasets. We find that the accuracy of our approach is influenced by the lexical similarity between utterances of different intents and the underlying distribution of data used to generate memberships. Results reveal that taking the underlying data distribution into account when generating memberships yields more consistent results in mapping and emulating binary memberships. Moreover, accounting for similarity between not only the intersections but also the string remainders yields the highest accuracy. \\

\noindent\textbf{Acknowledgements }This work\footnote{This is a preprint of the accepted manuscript: Geetanjali Bihani and Julia Taylor Rayz, Fuzzy Classification of Multi-intent Utterances, to be presented at NAFIPS 2021, whose proceedings will be published in \emph{Explainable AI and Other Applications of Fuzzy Techniques}, edited by Julia Taylor Rayz, Victor Raskin, Scott Dick, and Vladik Kreinovich, reproduced with permission of Springer Nature Switzerland AG. The final authenticated version will be available online at: (url tbd)} is partially supported by National Science Foundation 
grant number 1737591.




\bibliographystyle{spmpsci}
\bibliography{ref00.bib}

\end{document}